\documentclass[10pt,twocolumn,letterpaper]{article}
\usepackage{wacv}

\usepackage{graphicx}
\usepackage{comment}
\usepackage{amsmath}
\usepackage{amssymb}
\usepackage[square, comma, sort&compress, numbers]{natbib}
\usepackage{booktabs}
\usepackage{bm}
\usepackage{pifont}
\usepackage{amssymb}
\usepackage{amsmath}

\usepackage{ulem}
\usepackage{caption}

\usepackage[pagebackref,breaklinks,colorlinks]{hyperref}

\usepackage[capitalize]{cleveref}
\crefname{section}{Sec.}{Secs.}
\Crefname{section}{Section}{Sections}
\Crefname{table}{Table}{Tables}
\crefname{table}{Tab.}{Tabs.}

\def\wacvPaperID{1799} 
\def\confName{WACV}
\def\confYear{2025}

\begin{document}

\title{Structure-Aware Human Body Reshaping with \\ Adaptive Affinity-Graph Network}

\author{
Qiwen Deng$^{1*}$
\and
Yangcen Liu$^{1,2*}$
\and
Wen Li$^{1}$
\and
Guoqing Wang$^{1}$ 
\and
$^{1}$University of Electronic Science and Technology of China \\
$^{2}$Georgia Institute of Technology \\
\tt\small \{don2889632705,liwenbnu\}@gmail.com, yliu3735@gatech.edu, gqwang0420@hotmail.com
}

\twocolumn[{%
    \renewcommand\twocolumn[1][]{#1}%
    \maketitle
    \begin{center}
        \vspace{-6mm}
        \includegraphics[width=0.95\linewidth]{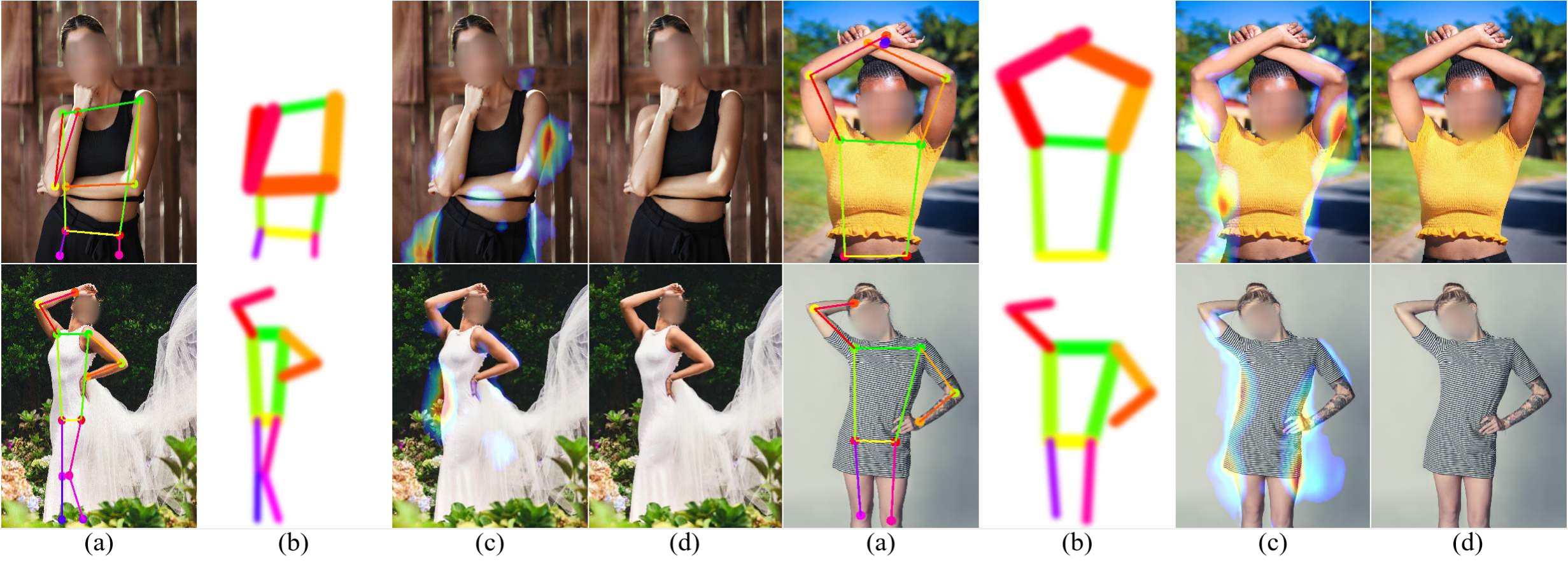}
        \captionof{figure}{Examples of the primary processes of our methods. (a) To reshape the human body, our Adaptive Affinity-Graph Network (AAGN) first extracts its skeleton map. (b) An affinity graph is constructed to regulate the consistency of human body parts. (c) The optical flow for warping is estimated. (d) Finally, the images are warped with the flow.}
        \label{fig:top}
    \end{center}%
    \vspace{-2mm}
}]

\maketitle

\renewcommand{\thefootnote}{\fnsymbol{footnote}}
\footnotetext[1]{Equal contribution.}

\begin{abstract}
Given a source portrait, the automatic human body reshaping task aims at editing it to an aesthetic body shape. 
As the technology has been widely used in media, several methods have been proposed mainly focusing on generating optical flow to warp the body shape. 
However, those previous works only consider the local transformation of different body parts (arms, torso, and legs), ignoring the global affinity, and limiting the capacity to ensure consistency and quality across the entire body. 
In this paper, we propose a novel \textbf{Adaptive Affinity-Graph Network} (\textbf{AAGN}), which extracts the global affinity between different body parts to enhance the quality of the generated optical flow. 
Specifically, our AAGN primarily introduces the following designs: 
(1) we propose an \textbf{Adaptive Affinity-Graph} (\textbf{AAG}) Block that leverages the characteristic of a fully connected graph. 
AAG represents different body parts as nodes in an adaptive fully connected graph and captures all the affinities between nodes to obtain a global affinity map. 
The design could better improve the consistency between body parts.
(2) Besides, for high-frequency details are crucial for photo aesthetics, a \textbf{Body Shape Discriminator} (\textbf{BSD}) is designed to extract information from both high-frequency and spatial domains. 
Particularly, an SRM filter is utilized to extract high-frequency details, which are combined with spatial features as input to the BSD. With this design, BSD guides the \textbf{Flow Generator} (\textbf{FG}) to pay attention to various fine details rather than rigid pixel-level fitting. 
Extensive experiments conducted on the BR-5K dataset demonstrate that our framework significantly enhances the aesthetic appeal of reshaped photos, surpassing all previous work to achieve state-of-the-art in all evaluation metrics. 
Our code is available at \href{https://github.com/Randle-Github/AGGN}{https://github.com/Randle-Github/AGGN}.

\end{abstract}

\section{Introduction}

Body reshaping aims to retouch a human body photo to make it more attractive, which is widely applied in media and other applications (e.g., mobile phones, advertisements, and apps). With the application of this technology, ordinary users can quickly obtain satisfying photos, while human editors can significantly enhance their work efficiency.

Several previous works have been proposed on this task. 
The body contour-based methods use non-rigid deformation~\cite{a1,a2} based on detected body contour. 
Its effectiveness is highly dependent on the accuracy of body contour detection, which means it cannot handle complex body poses.
A straightforward RGB-based approach is to generate reshaped photos directly~\cite{a16, a17}. 
However, with high-resolution and limited datasets, these methods often struggle with pixel-level information loss, such that the generated photos cannot fully restore the detailed characteristics of the input portraits. 
3D-based methods, such as NeRF \cite{mildenhall2021nerf}, have demonstrated their effectiveness across a wide range of tasks \cite{huang2022hdr, pianwan, mildenhall2022nerf}, including human body modeling \cite{jiang2022neuman, NEURIPS2021_65fc9fb4}. In particular, model-based approaches \cite{a3, a4, a5} construct adaptable 3D representations that can be dynamically adjusted to reshape the human body. However, these methods often rely on camera scanning, which imposes significant constraints on efficiency and scalability.
Recently, diffusion-based methods~\cite{personimagediffusion,disco} are designed for conditioned image generation, demonstrating extraordinary quality for the generated images.
Unfortunately, these methods commonly have high computation demands, and cannot be utilized by common users (e.g. on apps).

In this case, the flow-based method FBBR~\cite{fbbr} has been a reliable solution.
This approach utilizes a GAN-based model to generate optical flow based on human body photos, which is then utilized to warp the photos.
At first, this technique was primarily applied to the tasks of human faces~\cite{tal,atw} rather than body reshaping, as the previous models were unable to handle the complex structure of the human body. 
Due to the sparsity of optical flow, and avoiding RGB information leakage of the original photo, the optical flow-based methods achieve impressive results and low computation cost.
Despite outperforming all traditional methods in terms of both photo quality and efficiency, FBBR~\cite{fbbr} has two main weaknesses.
(1) It overlooks the global affinity of different body parts.
As FBBR simply trained the flow generator with local information of different body parts, the consistency of the whole body is ignored. 
(2) FBBR only employs a simple pixel-level training strategy in the spatial domain (e.g. skin color), making the model unable to extract more detailed information in the frequency domain (e.g. body contour). 
As the aesthetic appeal of the photo heavily relies on the high-frequency domain, human preference is strongly harmed.

In this study, we suppose that the deformation of a certain body part needs to take coordination with all other parts into consideration. 
Inspired by the structure of a fully connected graph, where each node is linked to all other nodes, we model different body parts as nodes within such a graph. The edges connecting these nodes represent the affinities between neighboring parts, allowing our model to effectively coordinate the depicted body regions. Besides, the edge weights signify the intensity of these affinities, reflecting their importance to photo aesthetics.
To achieve this, we propose the \textbf{Adaptive Affinity-Graph} (\textbf{AAG}) module, which computes mutual affinities between pairs of nodes in the fully connected body parts graph. 
By incorporating this module, the features extracted from each body part inherently involve contributions from all other parts, ensuring a feature map that captures global information. 
High-frequency features are crucial for enhancing photo aesthetics, but previous training strategies focused solely on the spatial domain, limiting the model's ability to capture fine details. To overcome this, we introduce the \textbf{Body Shape Discriminator} (\textbf{BSD}), which leverages both spatial and high-frequency domain features to identify relevant patterns in photos. We use SRM filters to extract high-frequency details, which are then combined with the original photos as input to BSD.
In addition to fitting target photos at the pixel level, the \textbf{Flow Generator} (\textbf{FG}) is designed to fit high-frequency features as well, aiming to fool the BSD. This approach allows FG to produce more vivid photos that closely match human designers' work, avoiding rigid distortions of specific body parts.
Figure~\ref{fig:pipeline} illustrates the basic process of our framework and the roles of different modules within it.

To summarize, our contribution is threefold:
\begin{itemize}
   \item We propose an Adaptive Affinity-Graph Block. It can capture global affinity among different body parts, boosting the usability of the flow generator.
   \item We design a novel adversarial training strategy with the Body Shape Discriminator, which is able to guide the Flow Generator to learn spatial and high-frequency domain features of photos, thereby aligning better with artificial editing results.
   \item Extensive comparative experiments have demonstrated that our method significantly surpasses all previous approaches in terms of different evaluation metrics.
\end{itemize}

\section{Related Works}

\subsection{Portrait Reshaping}
Reshaping human portraits has been widely used in social media and photography production. Numerous previous works have been proposed for this task.
Some early works~\cite{a1,a2} detect the contour of the human body, and then use non-rigid deformation based on the body contour to achieve the desired aesthetic effect.
Another series of methods~\cite{a3,a4,a5,3dreshaping} aims to create a 3D adaptable structure using a single portrait image. 
The structure can then be adjusted to alter body shape by manipulating parameters associated with body attributes.
However, these methods rely too heavily on the detection of contours and require an extremely complex artificial design.
Complex human gestures and inaccurate contour detection would experience severe performance degradation.
Also, most previous work are primarily used in tasks related to human faces~\cite{tal,atw}, for models were unable to handle the complex body structure.
Ren~\cite{a16} introduced a global-flow local-attention framework to reassemble source feature patches to the target. Tang~\cite{tangstructure} tackled consistency and coherency issues in video portrait reshaping. Prior to our work, FBBR~\cite{fbbr} was the only optical flow-based method for body shaping, incorporating a structure-aware framework with a skeleton map to enhance optical flow generation. In contrast, our method refines the output flow using an affinity graph between body parts, improving consistency and reducing overfitting.

\begin{figure*}[t]
  \centering
       \includegraphics[width=1.0\linewidth]{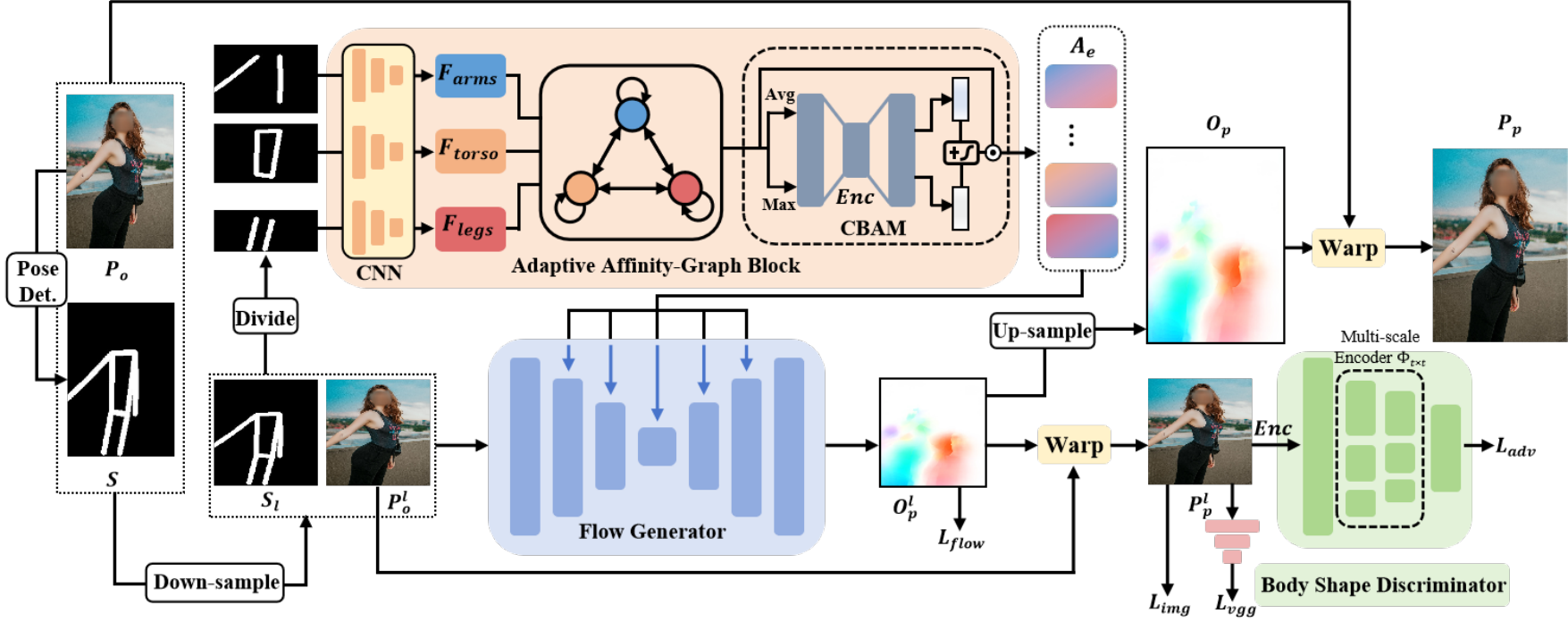}
       \caption{Overall pipeline of AAGN. The pipeline is mainly composed of Adaptive Affinity-Graph Block (AAG), Flow Generator (FG), and Body Shape Discriminator (BSD). First, the original photo and skeleton maps are downsampled. Then, in AAG, different body parts are separately encoded with CNN as an input of the Affinity-Graph. The body part features are reweighted through CBAM with the calculated affinity graph. In FG, conditioned by the output of AAG, optical flow for warping is estimated. The optical flow is then used to perform a warp operation on the downsampled original photo. In the training process, the BSD helps regulate the shape of the predicted photo to improve aesthetics.}
       \label{fig:pipeline}
       \vspace{-4mm}
\end{figure*}

\subsection{Pose-guided Image Generation}

Pose-guided Image Generation aims at generating portraits directly based on a human pose.
The problem has been studied extensively in recent years, especially with the unprecedented success of
GAN-based models~\cite{conditionedgan,poseguidedanimationfromsingle} for conditional image synthesis.
Then~\cite{unetshape} attempts to disentangle the appearance and the pose of person images using VAE-based design and a UNet-based skip-connection architecture.
Subsequently, some works~\cite{a16,liquidwarping} use flow-based deformation to transform the source information to improve pose alignment.
ADGAN~\cite{adgan} uses a texture encoder to extract style vectors for human body parts and gives them several residual blocks to synthesize the final image. 
Other methods~\cite{spgnet,pise,casd} make use of parsing maps to generate the final image.
CoCosNet~\cite{cocosnet,cocosnetv2} extracts dense correspondences between cross-domain images with attention-based operation. 
Recently, PIDM~\cite{personimagediffusion} is the first to propose a diffusion-based framework.
Disco~\cite{disco} is proposed for pose-driven video generation and significantly improves human motion quality with the diffusion model.
Although diffusion-based methods could improve the quality, their computational cost is not suitable for quantitative applications. The mainstream solution to our task is through flow warping.

\subsection{Optical Flow}

Optical flow estimation has been a longstanding research problem in computer vision~\cite{Determiningopticalflow,iteraitiveopticalflow,raft,pwcnet}. Its objective is to estimate the motion between consecutive video frames. 
As we recognize the source images to the target images as a transformation of motion, it’s worth noting that the key part of automatic portrait reshaping is also to predict the optical flow between two images.
Motion transfer~\cite{synthesizingimagesofhumans,posepersonimagegeneration,disentangledimagegeneration,deformablegans,poseguidedanimationfromsingle,pise} is a relevant task focusing on pose-guided human image generation. Monkey-Net~\cite{monkeynet} was the first to propose a flow-based method building backward motion flow from aligned keypoints to warp the source image feature to driving pose. 
Recent more advanced works~\cite{taojiale,xuborun,fomm,regionmm,yangcen1,yangcen2} mainly follow the optical flow-based strategy to process the problem as a flow estimation based on structure information for conditioning the image generation process and demonstrating better performance.
As for the automatic portrait reshaping task, ATW~\cite{atw}, FAL~\cite{tal}, and FBBR~\cite{fbbr} aim to predict the flow for body warping directly based on the estimated pose.
Our method is inspired by those methods to better extract the relationship between pose graph structure priors to the generated optical flow.

\section{Methods}
The overall network architecture of our approach is shown in Figure~\ref{fig:pipeline}. 
A high-resolution input photo $P_o$ and its skeleton map $S$ are downsampled and fed into the Flow Generator (FG). 
Then we organize the skeleton map into a fully connected graph structure, where different nodes of the graph represent different body parts.
A novel Adaptive Affinity-Graph (AAG) Block is proposed to capture the global affinity maps from the body parts graph. 
These maps are then embedded into the FG, enabling it to coordinate various areas throughout the body.
In the training stage, we directly warp the downsampled photo $P_o^l$ with generated flow $O_p^l$, obtaining low-resolution predicted photo $P_p^l$. 
The Body Shape Discriminator (BSD) determines whether $P_p^l$ and the target photo is manually retouched based on features in both spatial and high-frequency domain, guiding the FG to focus on these features. 
In the inference, generated flow $O_p^l$ is unsampled to the original size of $P_o$ and used to warp $P_o$, getting $P_p$. 

\subsection{Human Body Structure Construction}
We extract a skeleton map and organize different parts into a fully connected graph to construct the human body structure.
2D pose estimation methods are applied to predict the key points that can represent the distribution of the skeleton. By linking specific pairs of key points, we obtain a stack of skeleton maps $S = (s_1, s_2, ..., s_N)$ where $s_i$ is the $i$-th limb of the body in $S$.
Then, we divide different parts of the body into arms, torso, and legs and combine $s_i$ according to semantic separation into $S_{arms}$, $S_{torso}$ and $S_{legs}$. 
After downsampling these skeleton maps, to facilitate the feature analysis of the subsequent module, we use three convolutional blocks to extract features from different parts separately, obtaining $F_{arms}$, $F_{torso}$, $F_{legs}$. These three features serve as the nodes to construct a fully connected Affinity-Graph.

\subsection{Adaptive Affinity-Graph Block}
\label{sec:aag}
To maintain the body aesthetics, we hope that body parts are consistent with each other after reshaping. 
This means that when the Flow Generator (FG) is warping a certain body part, it is necessary to consider the affinities between this part and all other parts. 
All previous work ignores this issue and solely extracts the local information of different body parts, leading to a low capacity to coordinate the whole body.
To capture the global affinity, we propose the Adaptive Affinity-Graph Block (AAG) that constructs a fully connected graph structure. Taking advantage of the characteristic of this structure where each node is interconnected, AAG depicts different parts as nodes and computes the representation of each edge by capturing the affinities of neighbors. Due to the fully connected nature, the feature maps computed by AAG for each part include its affinities with all other body parts. In addition, we define the edge weights as the intensity of these affinities, which reflect their importance to photo aesthetics. By assigning weights to edges, AAG is able to recognize the contribution of different affinities to the photo aesthetics.

According to the design above, the process of the AAG encompasses two key stages: capturing affinity maps among body parts and assigning weights to these maps. In the initial stage, we classify the affinity among body parts into internal affinity within specific parts (\textit{e.g.}., left arm-right arm, left leg-right leg) and cross-affinity between different parts (\textit{e.g.}, arms-legs, arms-torso).
To achieve it, we apply self-attention and cross-attention mechanism to $F_i \in \mathbb{R}^{H \times W \times C}$ and $F_j \in \mathbb{R}^{H \times W \times C}$ $(i,j \in \{arms, legs, torso\})$, obtaining correlation map $A_{i-j} \in \mathbb{R}^{H \times W \times C} $. Firstly, we compute the operands of the attention mechanism:
\begin{equation}
    Q_{i-j} = \bm{\phi}_q^{i-j}(F_i), \xspace K_{i-j} = \bm{\phi}_k^{i-j}(F_j),\xspace V_{i-j} = \bm{\phi}_v^{i-j}(F_j),
\end{equation}
where $\bm{\phi}$ is 1 $\times$ 1 convolution kernel. 

Subsequently, we derive the correlation between regions of the $i$ part and the $j$ part by computing $W_{i-j}\in \mathbb{R}^{HW \times HW}$. 
To discern the pertinent region in the $j$-th part for the $i$-th part, we compute the affinity map $A_{i-j}$ as follows:
\begin{equation}
    W_{i-j} = \frac{Q_{i-j} \cdot K_{i-j}^T}{\sqrt{d_C}}, \quad \quad A_{i-j} = W_{i-j}^T \cdot V_{i-j},
\end{equation}
where $d_C$ is the number of channels. When $i=j$, the above operations represent the self-attention mechanism, and $A_{i-j}$ is an internal affinity within each body part. 
Conversely, when $i \neq j$, it illustrates the cross-attention mechanism, with $A_{i-j}$ denoting the cross affinity between different body parts.

After acquiring all the affinities that form the edges of the graph, our objective is to assign weights to these edges, reflecting the significance of specific affinities to the visual appeal of the photograph. 
Drawing inspiration from the channel-wise attention mechanism, which allocates weights to different channels, we initially employ a sharing encoder $\bm{Enc}_{i-j}$ to compress each affinity value into a single channel and concatenate them in the form of channels within $\bm{A} \in \mathbb{R}^{H \times W \times N^2}$, where $N$ represents the number of nodes:
\begin{equation}
    \bm{A} = Concat_{i,j}^N \bm{Enc}_{i-j}(\bm{A}_{i-j}).
\end{equation}

In this case, different affinities are in different channels of $\bm{A}$. After utilizing the channel-wise attention mechanism, we assign weights to affinities. Inspired by CBAM CBAM~\cite{a15}, we first compress the features from different channels: 
\begin{equation}
    A_{max}^c = MaxPool(A), \quad\quad A_{avg}^c = AvgPool(A).
\end{equation}

Though both Max-Pooling and Average-Pooling can compress the features from different channels, it has been proved that combining them can significantly improve the representation power of networks~\cite{a15}.
For capturing channel dependencies, we employ a shared CNN $\bm{Enc}_s$ to handle the condensed features. 
The resulting $W_c \in \mathbb{R}^{N^2}$ signifies a reassigned attention of affinities, representing the adaptive weights of the edges within the graph:
\begin{align}
    W_c= \sigma(\bm{Enc}_s(A_{max}^c) + \bm{Enc}_s(A_{avg}^c)).
\end{align}
For every channel of A, it corresponds to each weight in $W_c$.
By multiplying these weights with the channels, we assign different strengths to affinities as follows:
\begin{align}
    A_e = W_c \cdot A,
\end{align}
where $A_e\in \mathbb{R}^{H \times W \times N^2}$ represents the global affinities of different parts and the importance of a certain affinity to the photo aesthetics.
Finally, $\bm{A}_e$ will be embedded into the FG as a condition for generating optical flow.

\begin{figure}[t]
  \centering
       \includegraphics[width=0.7\linewidth]{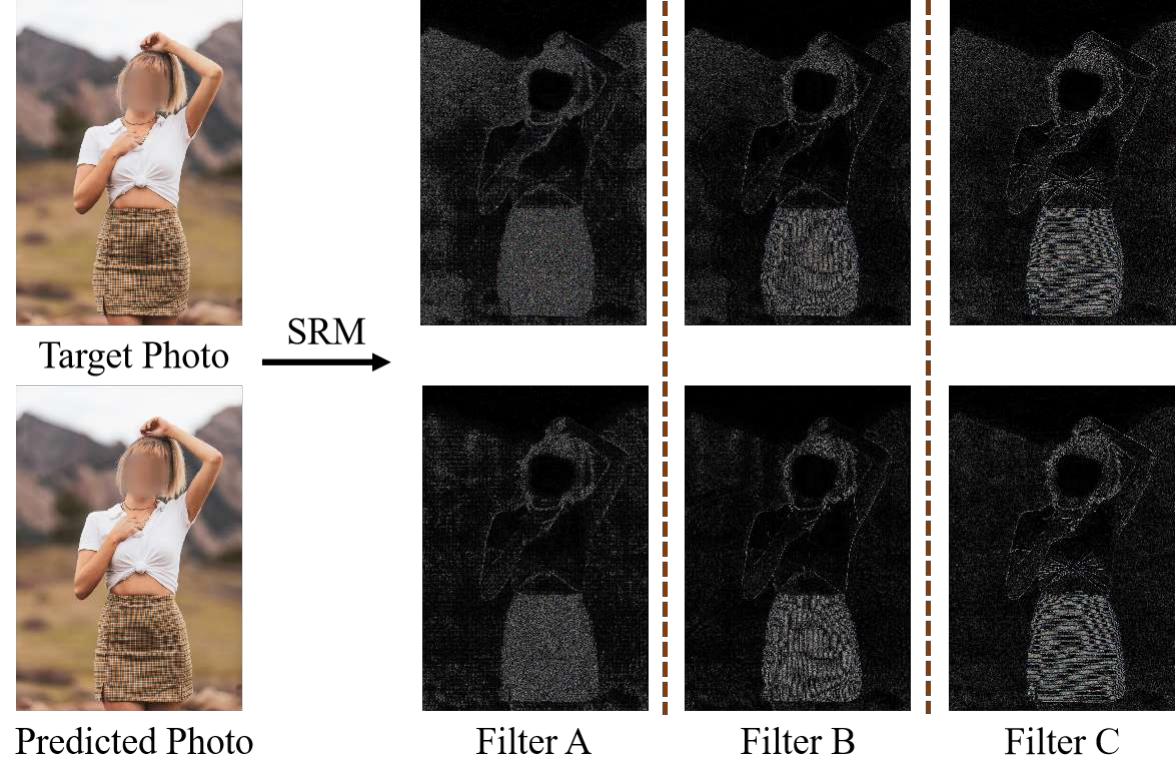}
       \caption{The SRM filters consist of three different filters. We use these three to filter the photo, getting high-frequency features. The Body Shape Discriminator can distinguish photos more accurately based on both these features.}
       \label{fig:SRM}
       \vspace{-4mm}
\end{figure}

\subsection{Merging Features}
We insert the global-aware affinity maps $\bm{A}_e$ into feature maps at different levels within the Flow Generator (FG), enabling it to extract richer information. We denote the original feature map of the $i_{th}$ layer of the FG as $\bm{FG}_i \in \mathbb{R}^{H_i \times W_i \times C_i}$.
Then a convolution-based encoder encodes $\bm{A}_e$ to $\bm{A}^e_i \in \mathbb{R}^{H_i \times W_i \times C_i}$ where $H_i$, $W_i$, $C_i$ represent the shape of feature maps in the $i$-th layer.
The embedding operation can be formulated as the sum of $\bm{A}^e_i$ and $\bm{FG}_i$:
\begin{align}
A^e_i = \bm{Enc}_i(A_e), \quad\quad FG_i^e = FG_i + A^e_i, \label{eq:combined}
\end{align}
With the output of AAG, the FG obtains predicted optical flow $O_p$ for body reshaping based on the original photo $P_o$ and skeleton maps $S$. After performing the warp operation $W(P_o,O_p)$, we get the predicted photo $P_p$. 

\subsection{Body Shape Discriminator}
In the training process, the Body Shape Discriminator (BSD) serves to discern whether the target photograph $P_t^l$ and the predicted photograph $P_p^l$ have been manipulated by a designer or generated by the Flow Generator (FG).

Considering that different areas of a photo may exhibit varying degrees of distortion, it’s essential to account for these variations during photo analysis. 
Therefore, we incorporate a multi-scale convolution module into the BSD to extract features at different scales. 
We denote the feature map of the $k_{th}$ layer of the BSD as $B_k$. Multi-scale feature extraction of the $k_{th}$ feature layer can be formulated as:
\begin{equation}
    B_k^{all} = \frac{(\bm{\phi}_{3\times3}(B_k) + \bm{\phi}_{5\times5}(B_k) + \bm{\phi}_{7\times7}(B_k))}{3},
\end{equation}
where $\bm{\phi}_{t \times t}$ is a ${t \times t}$ convolution kernel. 
For high-frequency features that are crucial to photo aesthetics, the BSD distinguishes the photos by the information from both the spatial and the high-frequency domain. 
To obtain the high-frequency features of photos, we use the SRM~\cite{srm} kernel to filter the target photo and the predicted photo, obtaining $F_{srm}$. The filtered results are shown in Figure \ref{fig:SRM}, highlighting the details (\textit{i.g.}, skin edge, fabric texture) that play an important role in photo aesthetics:
\begin{equation}
    F_{srm} = \bm{SRM}(P).
\end{equation}

After inputting a photo $P$ and its high-frequency features $F_{srm}$ into the BSD, it estimates the probability that the photo is produced by a human editor (ground truth) or the FG.
It could be utilized with an adversarial training strategy for optimizing the aesthetic of the generated portraits.

\subsection{Training Strategies}

We use PWC-Net\cite{pwcnet} to estimate the optical flow between the original photo ${P}_o$ and target photo ${P}_t$, obtaining a target flow ${O}_t$ to directly guide the Flow Generator (FG). Then we try to minimize the L2 distance between downsampled flows ${O}_t^l$ and ${O}_p^l$:
\begin{equation}
    L_{flow} = ||{O}_t^l - {O}_p^l||_2.
\end{equation}

For ${O}_t$ is not the accurate optical flow from ${P}_o$ to ${P}_t$, we also minimize the L2 distance of the downsampled predicted photo ${P}_p^l$ and target photo ${P}_t^l$ to fit the manually-retouched photo precisely:
\begin{equation}
    L_{img} = ||{P}_t^l - {P}_p^l||_2.
\end{equation}

Simply using pixel-wise loss functions is insufficient in handling the uncertainty inherent in recovering lost high-frequency details.
For the body reshaping task, the smoothness of the skin edge plays a crucial role in the aesthetic appeal of the photo. 
To alleviate this issue, we use a VGG loss, inspired by~\cite{a11}. We aim to minimize the distance between these two feature maps.
\begin{equation}
    L_{vgg} = ||\bm{V}_k({P}_t^l) - \bm{V}_k({P}_p^l)||_2,
\end{equation}
where $\bm{V}_k$ denotes the $k^{th}$ feature map of a pretrained lightweight VGG-16~\cite{vgg}.

In addition to the losses described so far, the adversarial loss is an important component of this game.
For the Body Shape Discriminator (BSD) can extract details in both the spatial domain and high-frequency domain of photos, the FG can align its behavior with designers by narrowing the gap between ${P}_p$ and ${P}_t$ in these details to effectively fool the Body Shape Discriminator $\bm{D}$. 
\begin{equation}
    \begin{split}
    L_{adv} = \sum_{n=1}^{N} - log(1 - \bm{D}({P}_p^l, F_{srm}^l)& 
    \\- log( \bm{D}({P}_t^l, F_{srm}^t)).&
    \end{split}
\end{equation}

Ultimately, the total loss $L^{BR}$ can be formulated as follows:
\begin{equation}
    L^{BR} = \lambda_{1}L_{flow} + \lambda_{2}L_{img} +  \lambda_{3}L_{vgg} + \lambda_{4}L_{adv},
\end{equation}
where $\lambda_{1}$, $\lambda_{2}$, $\lambda_{3}$ and $\lambda_{4}$ are the balancing factors.

\begin{figure*}[t]
  \centering
       \includegraphics[width=0.95\linewidth]{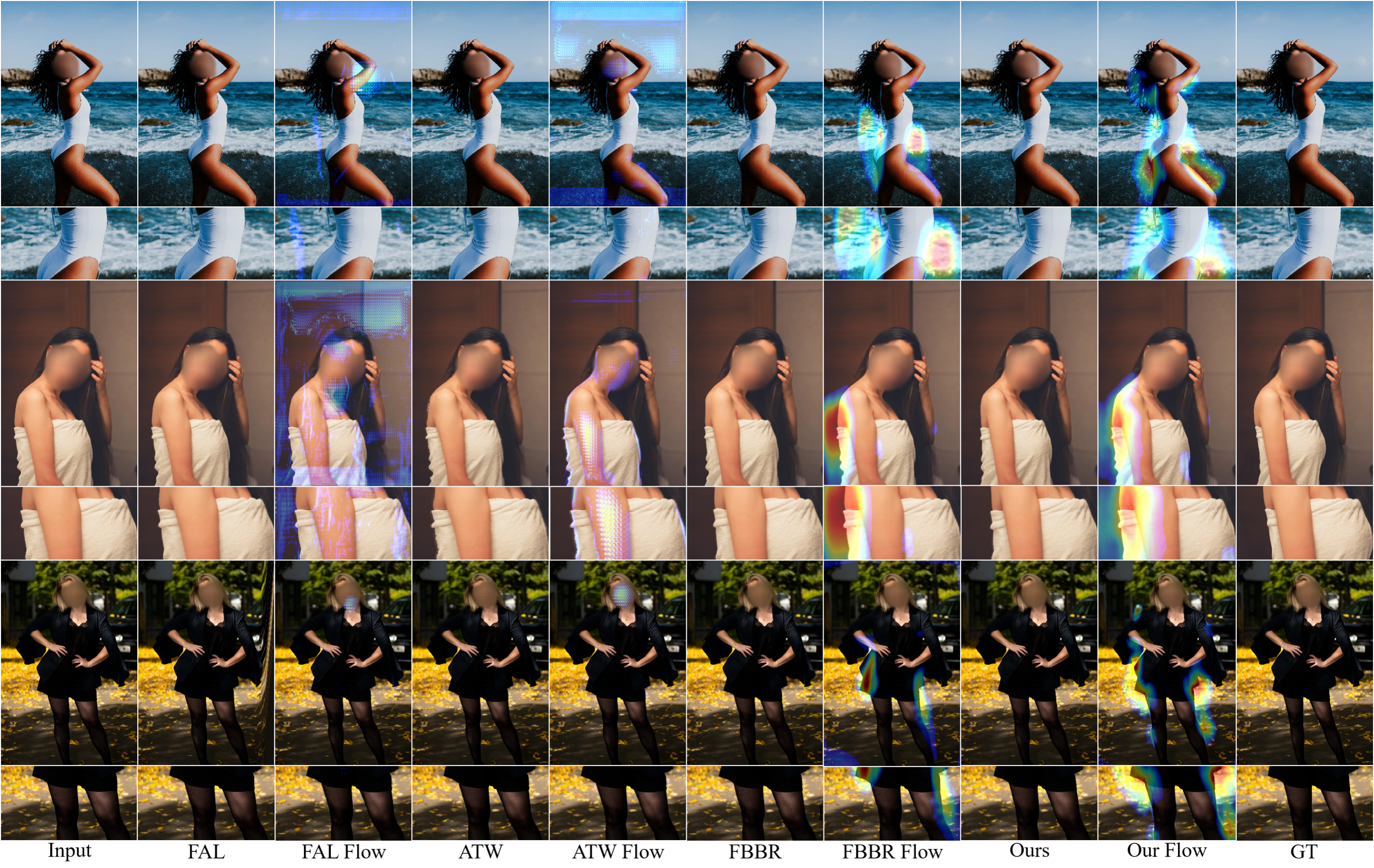}
       \caption{Visual comparisons among four optical flow-based methods, and the attention map of the estimated flow. Our method can produce high-resolution, believable, and consistent human body reshaping results. The attention map shows the more accurate grounding results of the torso, arms, and legs.
        }
       \label{fig:results}
       \vspace{-4mm}
\end{figure*}

\section{Experiments}
\subsection{Datasets and Settings}
\noindent\textbf{Datasets}. Due to privacy concerns surrounding our task, acquiring datasets for human body reshaping poses significant challenges. Therefore, we have chosen to evaluate our method using the BR-5K dataset~\cite{fbbr}, which stands as the sole publicly available dataset tailored for the human body reshaping task. Comprising 5000 high-quality individual portrait photos at a resolution of 2K, this dataset encompasses a diverse array of scenarios.
Each source portrait within BR-5K is accompanied by an artificially edited target photo, serving as a reference point for comparison. We proceed to randomly partition the dataset into a training set of 4500 photos and a test set of 500 photos.

\noindent\textbf{Implementation Details}. Skeleton maps $\bm{S}$ are plotted based on the key points predicted by the pose estimator. The 12 pairs of key points represent 12 different bones of the body. Connecting each pair of key points generates a map corresponding to the respective bone, and these 12 maps constitute the 12 channels of Skeleton maps $\bm{S}$. The weights in loss function are $\lambda_1 = 0.5$, $\lambda_2 = 0.5$, $\lambda_3 = 0.006$, $\lambda_4 = 0.001$. We train our framework using Adam optimizer with a learning rate of 1e-5 and batch size of 8.

\textbf{Evaluation Metrics}. 
Here we operate two kinds of comparison. 
(1) For quantitative evaluation, we employ three metrics: SSIM, PSNR, and LPIPS, to gauge the disparity between the predicted reshaped photo $\bm{P}_p$ and the target photo $\bm{P}_t$. 
PSNR assesses similarity by quantifying the mean squared error between the two images. 
SSIM, on the other hand, evaluates similarity by taking into account both structural information and perceptual phenomena inherent in the photos. 
LPIPS calculates similarity by comparing the activations of the two images using a predetermined neural network.
(2) For human preference assessment, we conduct a user study on 40 generated results produced by four flow-based methods.

\subsection{Comparison with Existing Approaches}
We only compared our method with available 2D-based methods as 3D model-based body reshaping approaches typically require significant user assistance or additional sensors, making them impractical for real-world applications.

There are five methods available for comparison: GFLA~\cite{a16}, pix2pixHD~\cite{a17}, FAL~\cite{tal}, ATW~\cite{atw}, FBBR~\cite{fbbr}. 
GFLA and pix2pixHD are traditional lightweight conditional image generation frameworks that were not specifically designed for human body reshaping. Therefore, we will compare these methods only in quantitative evaluations.
FAL and ATW, are optical flow-based methods originally designed for face-related tasks. 
FAL, designed for detecting face manipulations with local warping field prediction, is repurposed in our study to estimate target optical flow for body reshaping. 
Similarly, ATW, initially for facial expression animation, is adapted to generate optical flow for body reshaping. As noted in~\cite{fbbr}, the source photo and skeleton maps can serve as conditioning factors for generating the target photo.

\begin{table}[h]
    \centering
    \resizebox{0.7\linewidth}{!}{
    \begin{tabular}{l|cccc}
    \toprule
    Method &SSIM $\uparrow$ &PSNR $\uparrow$ &LPIPS $\downarrow$ \\
    \midrule 
    GFLA &0.6649 &21.4796 &0.6136 \\
    pix2pixHD &0.7271 &21.8381 &0.2800\\
    FAL &0.8261 &24.1841 &0.0837 \\
    ATW &0.8316 &24.6332 &0.0805 \\
    FBBR &0.8354 &24.7924 &0.0777 \\
    \midrule
    Ours &\textbf{0.8427} &\textbf{26.4100} &\textbf{0.0643} \\
    \bottomrule
    \end{tabular}}
    \caption{Quantitative comparison on BR-5K dataset for all six methods. The results are calculated as an average of 500 test photos and their reference artificial edited photos.}
    \label{tab:results}
    \vspace{-3mm}
\end{table}

\begin{table}[h]
    \centering
    \resizebox{0.4\linewidth}{!}{
    \begin{tabular}{l|c}
    \toprule
    Method & Preference $\uparrow$ \\
    \midrule
    FAL & $5.2\%$ \\
    ATW & $10.9\%$ \\
    FBBR & $32.1\%$ \\
    \midrule
    Ours & \textbf{51.8$\%$} \\
    \bottomrule
    \end{tabular}
    }
    \caption{User study for human preference of the four flow-based methods. User preferences favor our approach.}
    \label{tab:userstudy}
    \vspace{-3mm}
\end{table}

FBBR~\cite{fbbr} stands as the sole model explicitly tailored for the human body reshaping task. With the input of skeleton maps, their model achieves state-of-the-art performance. These methods are trained using the same settings as our method and evaluate their usability.

\noindent\textbf{Quantitative Comparison}. We employ these methods to reshape 500 photos from the test set and compute the four quantitative metrics between the predicted reshaped photos and target photos. The quantitative results are presented in Table~\ref{tab:results}.
From the table, it's evident that the two RGB-based methods, GFLA, and pix2pixHD, fail to produce satisfactory results, falling notably below the subsequent flow-based methods across all three metrics. We hypothesize that these methods suffer from RGB information leakage, leading to inadequate preservation of pixel details from the source image.
Among the four flow-based methods, the results of FBBR notably outperform those of FAL and ATW, primarily attributed to the incorporation of body structure information.

However, our AAGN surpasses all the aforementioned methods, reaching state-of-the-art performance. Specifically, we outperform FBBR by 0.0073, 1.6176, and 0.0134 in terms of SSIM, PSNR, and LPIPS, respectively.
While previous works can be adapted to human body reshaping tasks, the absence of modules for extracting body structure information constrains the effectiveness of these models.

\noindent\textbf{Human Preference Comparison}. 
Due to the poor performance of RGB-based methods in quantitative metrics, we focus exclusively on evaluating four optical flow-based methods in our user study. Figure~\ref{fig:results} illustrates several examples, including the torso, arms, and legs. We visualize the edited portraits and estimated optical flow for all four methods.
For FAL and ATW, it is difficult to identify the specific areas they edited. FAL is designed for detecting and reversing facial warping manipulations, while ATW generates high-quality facial expression animations through motion field prediction, making both methods unsuitable for human pose editing.
In contrast, our AAGN better targets all necessary body parts for transformation compared to FBBR. The visualizations confirm that our affinity graph effectively captures the correlations between body parts.

\begin{table}[t]
    \centering 
    \resizebox{0.85\linewidth}{!}{
    \begin{tabular}{l | c c c | c c c} 
        \toprule 
        Exp & FG & AG & BSD &SSIM $\uparrow$ &PSNR $\uparrow$ &LPIPS $\downarrow$ \\
        \midrule
        1 & \checkmark & \ding{55}& \ding{55} & 0.8340 & 26.20 & 0.0664 \\
        2 & \checkmark & \checkmark & \ding{55} & 0.8402 & 26.34 & 0.0655 \\
        3 & \checkmark &\ding{55} &\checkmark & 0.8382 & 26.33 & 0.0660  \\
        \midrule
        4 & \checkmark & \checkmark &\checkmark & \textbf{0.8427} &\textbf{26.41} &\textbf{0.0643} \\
        \bottomrule 
    \end{tabular}
    }
    \caption{Ablation study on the Flow Generator (FG), the Affinity-Graph (AG) block. and Body Shape Discriminator (BSD).}
    \vspace{-3mm}
    \label{tab:ablation}
\end{table}

We conducted a user study with 20 participants, each presented with 40 randomly selected questions from a pool of 40 samples. In each question, participants viewed a GIF showing the transformation from the source image to the target image for each method and then selected their preferred outcome based on visual quality.
As illustrated in Table~\ref{tab:userstudy}, our method garners the majority preference (59$\%$ of all participants), significantly surpassing the previous state-of-the-art method FBRR~\cite{fbbr}. These findings align with quantitative metrics, underscoring our method's advancement in terms of human preference.

\subsection{Ablation Study}
To validate the effectiveness of different modules in our model, we commence with an ablation study of modules.

\noindent\textbf{Ablation Study on Different Block}. Our Adaptive Affinity-Graph Network (AAGN) mainly consists of three main components: the Flow Generator (FG), the Affinity-Graph Block (AG), and the Body Shape Discriminator (BSD).
Table~\ref{tab:ablation} presents the results.
The baseline model solely possesses the Flow Generator (FG), and the results are presented in Exp. 1.
Then, we only incorporate Affinity Graph Block (AG) and Body Shape Discriminator (BSD) with FG in Exp. 2, 3. 
We observe significant improvements with our modules: AG achieves 0.0062, 0.14, and 0.0009, while BSD yields 0.0042, 0.11, and 0.0004. Combining AG and BSD further enhances performance to 0.0087, 0.21, and 0.0021, demonstrating their effectiveness in capturing global body part affinities and high-frequency details. 
First, joint training with object detection improves action recognition, which offline detectors cannot achieve. Second, incorporating interaction object information is essential for more accurate interaction modeling.

\begin{figure}[t]
  \centering
       \includegraphics[width=1.\linewidth]{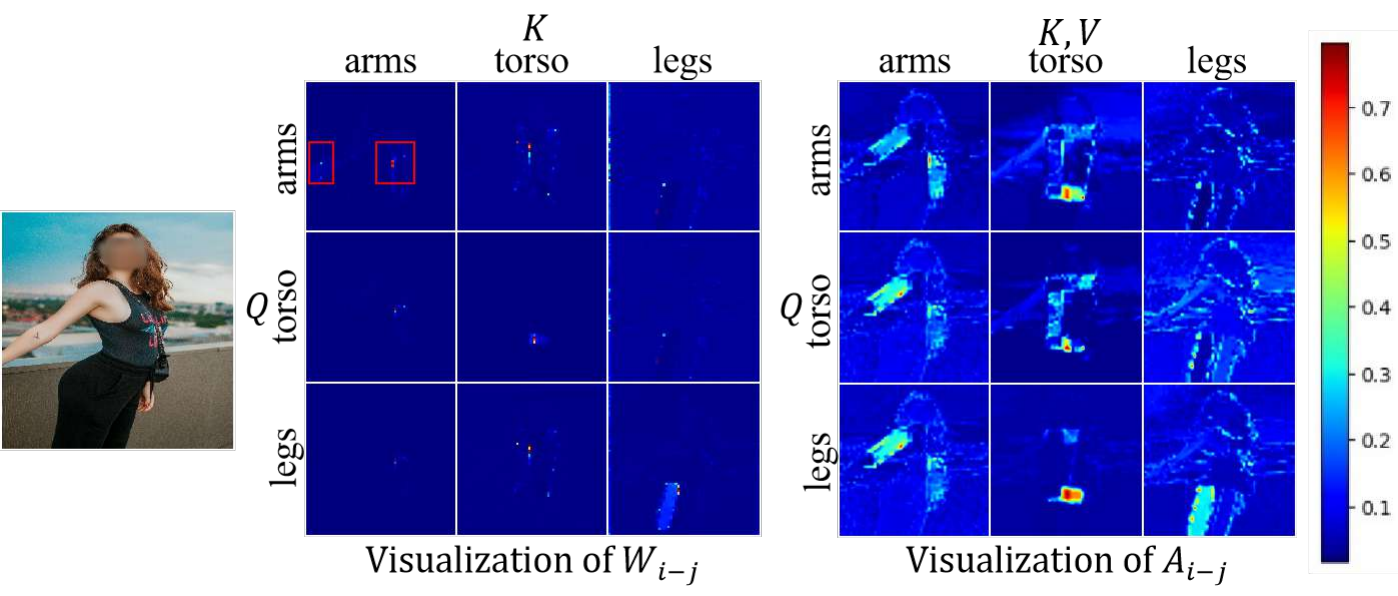}
       \caption{An example of visualization of the $W$ and $A$ in Adaptive Affinity-Graph Block. The \textcolor{red}{red bounding boxes} marks a sample of high attention region in $W_{i-j}$ and $A_{i-j}$.}
       \label{figs:9x9attention}
       \vspace{-4mm}
\end{figure}

\begin{figure}[t]
  \centering
       \includegraphics[width=1.0\linewidth]{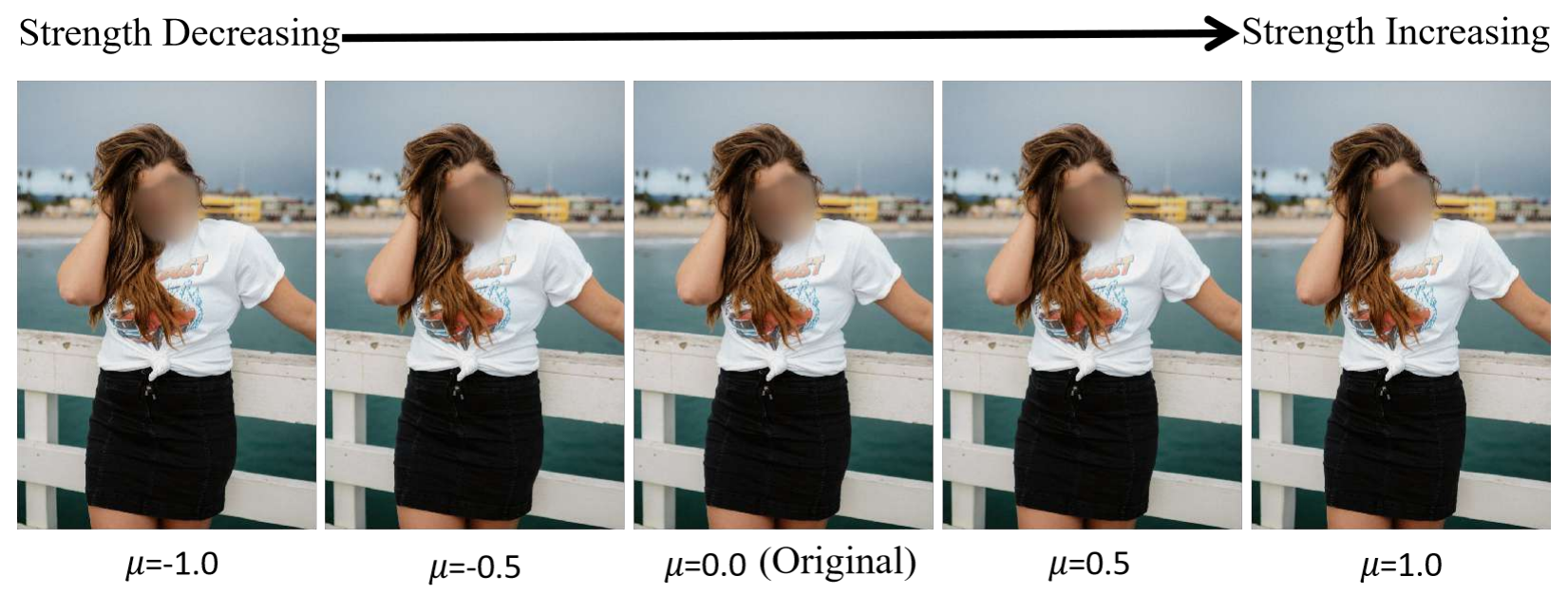}
       \caption{Our method provides continuous controls on body reshaping by adjusting $\mu$ in the warping procedure.}
       \label{figs:strength}
       \vspace{-4mm}
\end{figure}

\begin{figure}[t]
  \centering
       \includegraphics[width=1.0\linewidth]{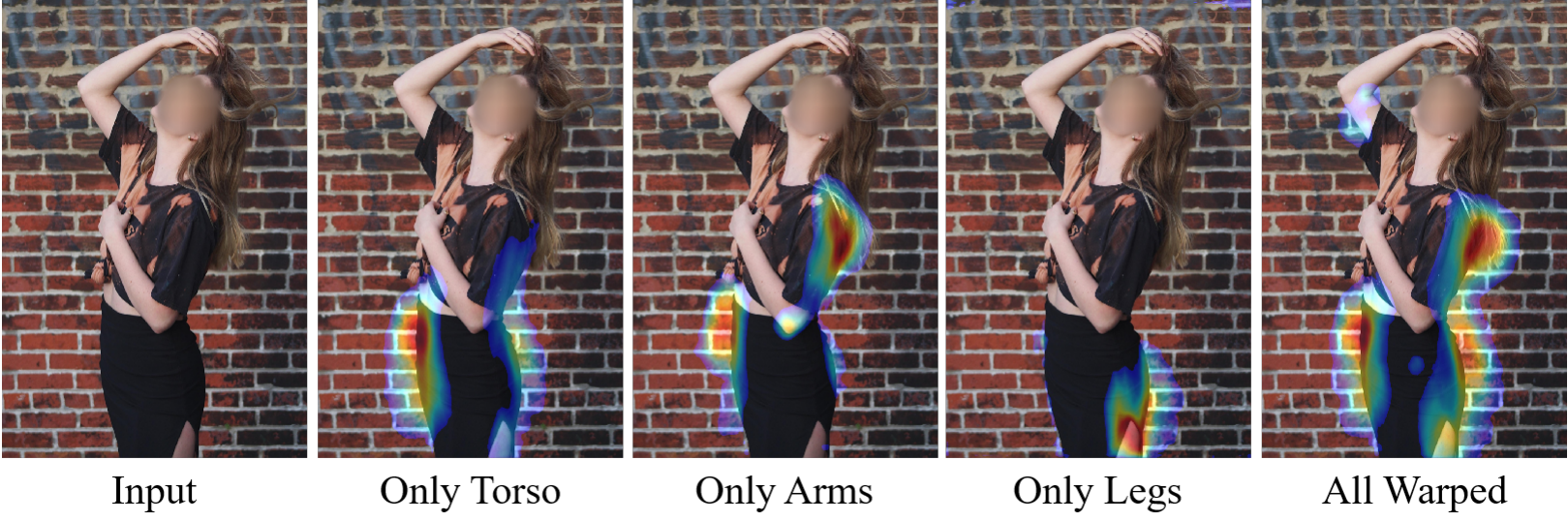}
       \caption{Our method provides local edit controls on body reshaping by adjusting the input skeleton map.}
       \label{figs:local}
       \vspace{-5mm}
\end{figure}

\noindent\textbf{Attention in Affinity-Graph}. As visualized in Fig~\ref{figs:9x9attention}, we make an example of the attention map of $W_{i-j}$ and $A_{i-j}$ (averaged by all channels) in Sec~\ref{sec:aag}.
For $W_{i-j}$, the attention values mostly concentrate on a few key regions.
We observe that the self-attention on the diagonal concentrates on their regions (for arms to arms, high-attention regions are on two arms).
As for cross-attention, the values are mainly high in the adjacent or relevant regions (for arms to torso, high-attention regions are primarily on the shoulder).
Then, for $A_{i-j}$, the attention map of $V$ is re-assigned by the affinity-graph weight.
The visualization results prove that our Adaptive Affinity-Graph can better adjust the correlation between different body parts.

\subsection{Applications}

In this section, we will show several user-related cases in real applications.

\noindent\textbf{Reshaping Strength Adaptation}.
One of the main advantages of optical flow-based methods is that they keep the users in control and can adjust the reshaping strength. For the users, with estimated optical flow, they could adjust the strength parameter as shown in Figure~\ref{figs:strength}. The strength parameter $\mu$ can be set from -1 to 1. A larger $|\mu|$ will introduce a bigger
change in body shape, while the positive/negative sign of $\mu$ controls whether to lose or gain weight.

\noindent\textbf{Local Editing}. 
Compared with previous work, a strength of our method is its ability to do local editing.
Unlike previous methods, our model contains an Affinity-Graph to regulate the consistency between different body parts.
To only reshape certain body parts to satisfy the users' demands, it is controllable to manipulate the local editing through the input skeleton map.
The results are shown in Figure~\ref{figs:local}.

\section{Conclusion}

In this paper, we introduce an Adaptive Affinity-Graph Network for human body reshaping, achieving state-of-the-art results across all evaluation metrics. Our method includes the Flow Generator (FG), Adaptive Affinity-Graph Block (AAG), and Body Shape Discriminator (BSD). We show that using skeletons as structural priors for AAG enhances image consistency by leveraging body part affinities. Additionally, the BSD’s adversarial training strategy regulates portrait aesthetics using SRM filters. Experiments on our BR-5K dataset confirm our method’s superior visual performance, controllability, and efficiency.

\setcounter{section}{0}

\renewcommand{\thesection}{\Roman{section}}
\section*{Supplementary Material}

\section{Global Affinity}
\begin{figure}
  \centering
       \includegraphics[width=1.0\linewidth]{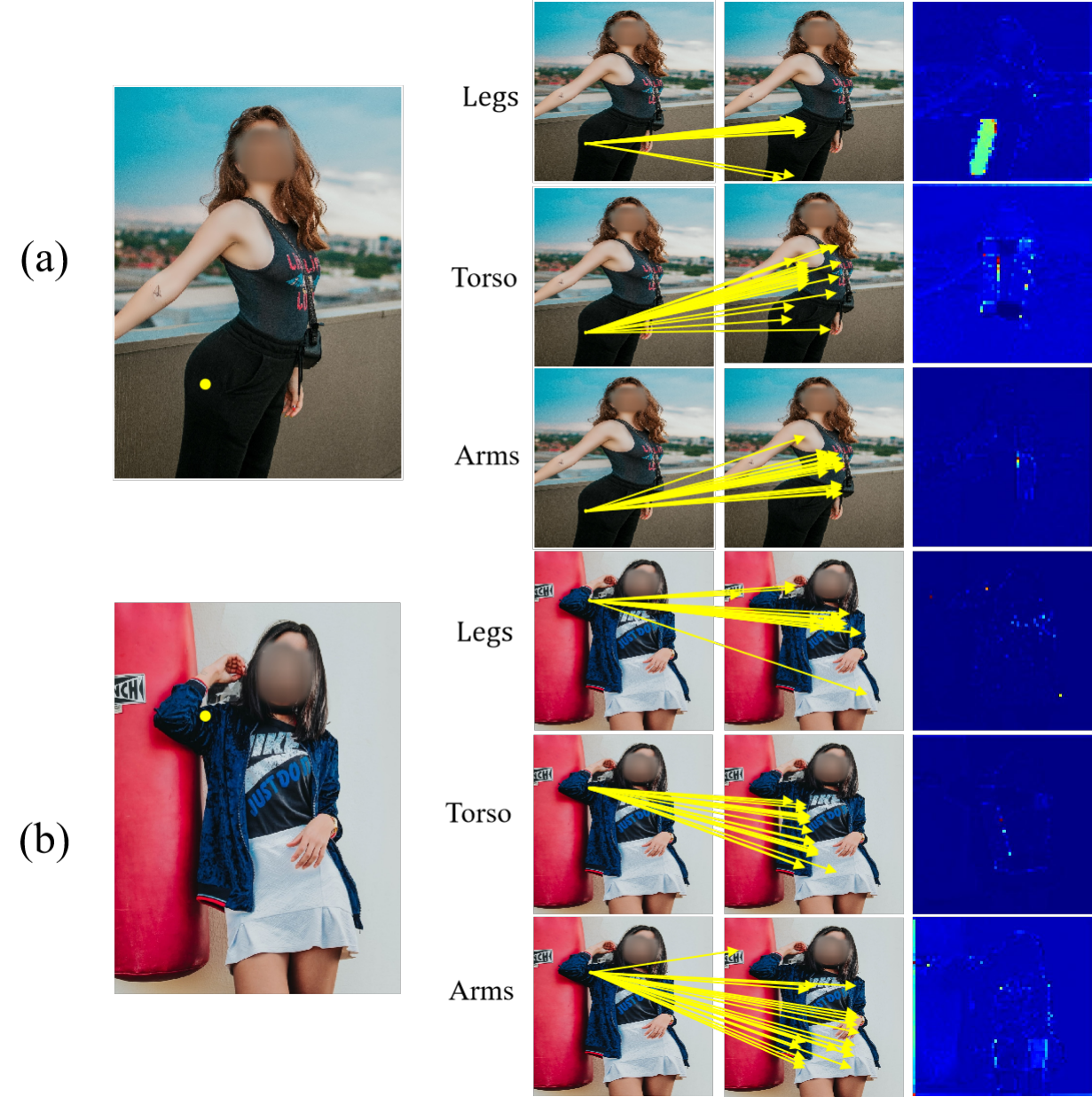}
       \caption{The visualization of $W$ queried by a single point. (a) shows the 20 highest activated points of a sample point in the legs, and (b) a sample point in the arms.}
       \label{fig:sup2}
\end{figure}

For a specific body part, ensuring consistency with other parts involves extracting affinity across the portraits in the Adaptive Affinity-Graph Block (AAG).
As illustrated in Figure~\ref{fig:sup2}, we visualize the affinity map $W$ for two single points, separately in arms and legs.
In (a), we visualize the activated attention map for legs, torso, and arms, with a query point in the legs, and we mark the top 20 activated points.
In (b) we visualize the results activated by a query point from the arms.
The body parts features are $F_{i}\in \mathbb{R}^{H \times W \times C}, i \in \{torso, legs, arms\}$.

In (b) we let $q$ represent the query point in $F_{arms}$ and $k$ represent a key point in $F_{torso}$. 
The attention map of point $q$ to $F_{torso}$ is computed as follow:
\begin{equation}
    W_{q,k} = \frac{1}{C}\sum_{c=1}^{C} Q_q^c * K_k^c,
\end{equation}
where $Q_q^c$ represents query value ($Q=\Phi_Q^{arms,torso}(F_{arms})$) of point $q$ in channel $c$ and $K_k^c$ represents the key value ($K=\Phi_K^{arms,torso}(F_{torso})$) of point $k$. 
The higher value of $W_{i,j}$ indicates the stronger affinity between point $q$ in the arms and point $k$ in the torso.

Our observations indicate that our model effectively directs attention to primary areas with affinity, thereby preserving photo aesthetics.
For example, in Figure~\ref{fig:sup2}(a), the attention map to torso,
points with high activation scores accumulated on the back (left) side of the torso, indicative of a strong association with the leg region.
Leveraging the Adaptive Affinity-Graph Block (AAG), which extracts affinity between each pair of body parts to obtain global affinity, our model achieves consistency across the entire body of the generated image.

\section{Ablation Study}
\noindent\textbf{Ablation Study on Convolutional Block Attention Module}
The Adaptive Affinity-Graph Block (AAG) is composed of two primary components: a fully connected graph network, wherein each node corresponds to a distinct body part for affinity extraction, and the Convolutional Block Attention Module (CBAM).
In our method, we organize affinities into the channel of $A$ and use a Convolutional Block Attention Module (CBAM) following~\cite{a15} to re-assign and refine the weights to these affinities utilizing the inductive bias of both average pooling and max pooling.

\begin{figure*}[t]
  \centering
       \includegraphics[width=0.85\linewidth]{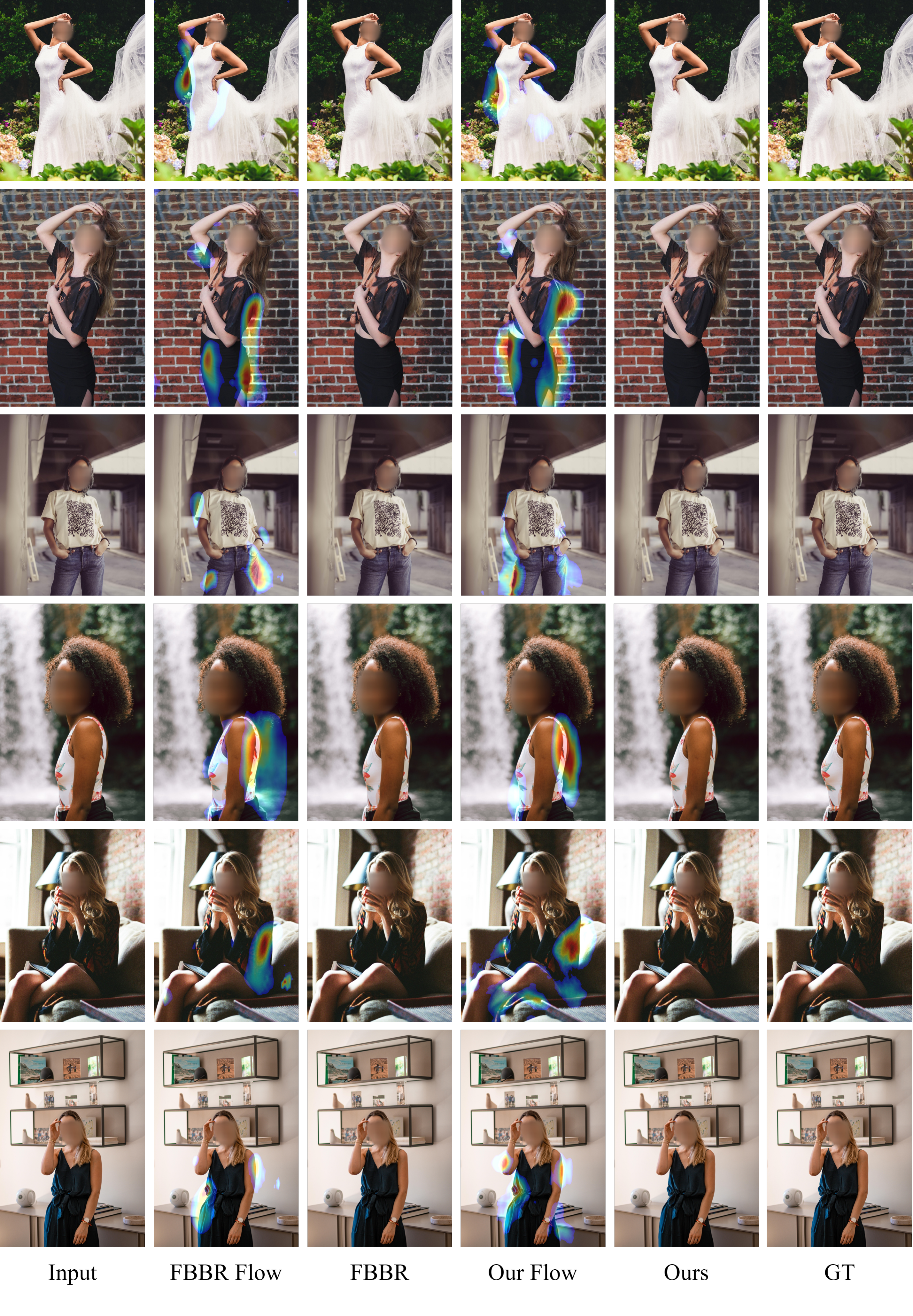}
       \caption{The qualitative results of our method and FBBR.}
       \label{fig:sup1}
\end{figure*}

The ablation study with or without the CBAM is shown in Table~\ref{tab:result}.
An improvement is evident across all three metrics.  
The results prove that while acquiring the global affinity, re-assigning and reigning the attention score to different body parts with CBAM is necessary for the output affinity map $A$.

\begin{table}[h]
    \centering
    \resizebox{0.85\linewidth}{!}{
    \begin{tabular}{l|cccc}
    \toprule
    Method &SSIM $\uparrow$ &PSNR $\uparrow$ &LPIPS $\downarrow$ \\
    \midrule 
    without CBAM &0.8398 &25.8853 &0.0679 \\
    with CBAM &\textbf{0.8427} &\textbf{26.4100} &\textbf{0.0643} \\
    \bottomrule
    \end{tabular}
    }
    \caption{Ablation study of Convolutional Block Attention Module (CBAM). The results are evaluated when Adaptive Affinity-Graph Block (AAG) is with or without CBAM design.}
    \label{tab:result}
\end{table}

\noindent\textbf{Reliance on pose estimation.} 
Our method requires skeleton maps as input. However, this reliance is acceptable for the following reasons: 
(1) A lower-bound analysis in Table~\ref{tab:lowerbound} indicates performance degradation in the extreme case where the input pose is masked. 
Despite this, the lower-bound performance still surpasses that of RGB-based methods. 
(2) Pose estimation is not the key point of our paper. We use the pose estimator aligned with FBBR for fair comparison, though more advanced pose estimation methods~\cite{humanposenew,unihcp} could further improve the results.

\noindent\textbf{Ablation study of flow generator.} The effectiveness of VGG loss is demonstrated in Table~\ref{tab:ablations}. We observed that without $L_{VGG}$, the performance in LPIPS is even better. We believe that $L_{VGG}$ helps with high-frequency information extraction, but might lead to extra flow blur in those irrelevant areas.

\begin{table}[ht]
    \vspace{-1em}
    \centering 
    \resizebox{0.85\linewidth}{!}{
    \begin{tabular}{l | c c | c c c} 
        \toprule 
        Exp & $L_{VGG}$ & $L_{img}$ &SSIM $\uparrow$ &PSNR $\uparrow$ &LPIPS $\downarrow$ \\
        \midrule
        1 & \ding{55} & \checkmark & 0.8351 & 26.27 & \textbf{0.0662} \\
        2 & \checkmark & \ding{55} & 0.8346 & 25.97 & 0.0719 \\
        \midrule
        3 & \checkmark & \checkmark & \textbf{0.8365} &\textbf{26.29} &0.0669 \\
        \bottomrule 
    \end{tabular}
    }
    \caption{Ablation study on Flow Generator.}
    \label{tab:ablations}
\end{table} 

\section{More Visualization Results}
\noindent\textbf{More Visualization}
We show more qualitative comparison results in Figure~\ref{fig:sup1}. Here we only compare our method with FBBR~\cite{fbbr} as shown in Figure~\ref{fig:sup1}. 
Our approach can produce more consistent and visually pleasing body-reshaping results.
We can observe that FBBR tends to edit a particular part individually rather than considering all parts collectively.

\noindent\textbf{Dynamic Visualization}
In this section, we present a package of 8 sample GIFs included in the supplementary material. We exclusively compare our model with the previous state-of-the-art method, FBBR. Our aim is to provide a clearer demonstration of the application process. The results highlight the advantages of our method and emphasize that optical flow-based methods excel at preserving the background of the original image. Optical flow primarily transforms specific areas, making it more suitable for our task. Conversely, recent trends in image-to-image transformation or relevant downstream tasks, such as motion transfer, have leaned towards employing diffusion models.

\begin{figure}
  \centering
       \includegraphics[width=1.0\linewidth]{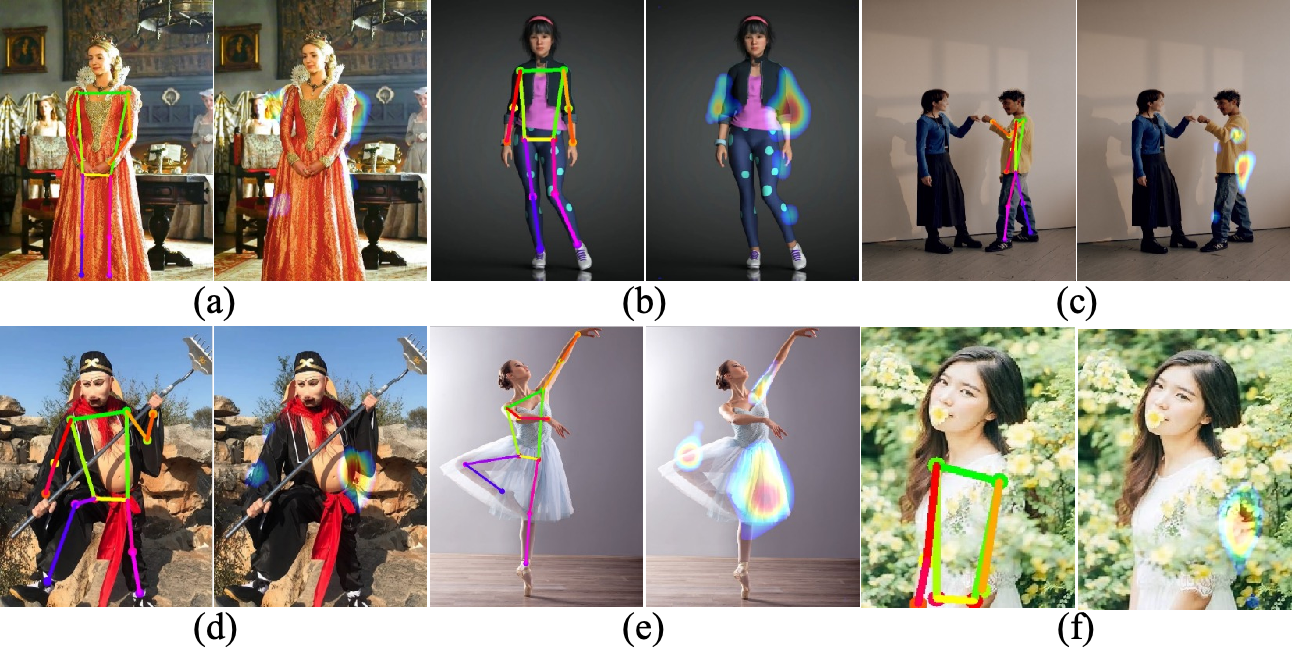}
       \caption{Results of multiple types of open-domain images. (a) Complex clothes. (b) Digital human. (c) Multiple persons. (d) Fantasy characters. (e) Complicated pose. (f) Occlusions.}
       \label{fig:opendomain}
\end{figure}


\begin{table}[h]
    \centering 
    \resizebox{0.8\linewidth}{!}{
    \begin{tabular}{l | c c c} 
        \toprule 
        Exp & SSIM $\uparrow$ &PSNR $\uparrow$ &LPIPS $\downarrow$ \\
        \midrule
        w/o pose & 0.8132 & 23.9434 & 0.0892 \\
        w/ pose & \textbf{0.8427} & \textbf{26.4100} & \textbf{0.0643} \\
        \bottomrule 
    \end{tabular}
    }
    \caption{Lower-bound analysis without pose input.}
    \label{tab:lowerbound}
\end{table}

\noindent\textbf{Calculation and structure differences compared to FBRR.}

The results of the calculation and parameter scale comparison are shown in Table~\ref{tab:fps}. 
Our AAGN improves performance while remaining lightweight compared to previous work, with only a 0.11ms FPS decrease and an additional 0.2M parameters. 
Our Flow Generator (FG) retains the same structure as in FBBR but omits the Structure Affinity 
\begin{table}[h]
    \centering
    \resizebox{0.75\linewidth}{!}{
    \begin{tabular}{l|cc|c}
    \toprule
    Exp & FBBR & FG & AAGN \\
    \midrule 
    FPS (ms) & 3.75 & 3.65 & 3.54 \\
    Param (M) & 6.7 & 6.6 & 6.9 \\
    \bottomrule
    \end{tabular}
    }
    \caption{Quantitative comparison of FPS (Frame Per Second) and Parameter Scale of processing images (including pose detector) on a single RTX 3090.}
    \label{tab:fps}
\end{table} 
Self-attention module.

\noindent\textbf{Visualization on open-world datasets.}
Testing the methods on different datasets is essential. However, similar to previous work, the BR-5K dataset is the only publicly available dataset. 
Despite concerns about generalizability, our method, trained on the limited BR-5K dataset, still delivers remarkable results on open-domain images. 
We visualize six types of open-domain images downloaded from the web, as shown in Figure~\ref{fig:opendomain}.
The visualization results still maintain high quality.

\small
\bibliographystyle{ieee_fullname}
\bibliography{main}

\end{document}